\documentclass[conference]{IEEEtran}
\IEEEoverridecommandlockouts

\usepackage{cite}
\usepackage{amsmath,amssymb,amsfonts}
\usepackage{algorithmic}
\usepackage{graphicx}
\usepackage{textcomp}
\usepackage{xcolor}
\usepackage{bm}
\usepackage{multirow}
\usepackage{colortbl}
\usepackage{floatrow}
\usepackage{makecell}
\usepackage{hyperref}

\hypersetup{urlcolor = blue}

\def\BibTeX{{\rm B\kern-.05em{\sc i\kern-.025em b}\kern-.08em
    T\kern-.1667em\lower.7ex\hbox{E}\kern-.125emX}}
\begin{document}

\def\eg{\emph{e.g.}} 
\def\Eg{\emph{E.g.}}
\def\ie{\emph{i.e.}} 
\def\Ie{\emph{I.e.}}
\def\cf{\emph{cf.}} 
\def\Cf{\emph{Cf.}}
\def\etc{\emph{etc}} 
\def\vs{\emph{vs.}}
\def\wrt{w.r.t.} 
\def\dof{d.o.f.}
\def\iid{i.i.d.} 
\def\wolog{w.l.o.g.}
\def\etal{\emph{et al.}}
\makeatother

\title{
    Unified Arbitrary-Time Video Frame Interpolation and Prediction
}

\author{
\IEEEauthorblockN{Xin Jin}
\IEEEauthorblockA{\textit{{\small Samsung \!Electronics \!(China) \!R\&D \!Centre}} \\
Nanjing, China \\
xin.jin@samsung.com}
\and
\IEEEauthorblockN{Longhai Wu}
\IEEEauthorblockA{\textit{{\small Samsung \!Electronics \!(China) \!R\&D \!Centre}} \\
Nanjing, China \\
longhai.wu@samsung.com}
\and
\IEEEauthorblockN{Jie Chen}
\IEEEauthorblockA{\textit{{\small Samsung \!Electronics \!(China) \!R\&D \!Centre }} \\
Nanjing, China \\
ada.chen@samsung.com}
\and
\IEEEauthorblockN{\quad \quad \quad \quad \quad \quad \quad \quad \quad Ilhyun Cho}
\IEEEauthorblockA{\textit{\quad \quad \quad \quad \quad \quad \quad \quad \quad Samsung Electronics}\\
\quad \quad \quad \quad \quad \quad \quad \quad \quad Suwon, Korea \\
\quad \quad \quad \quad \quad \quad \quad \quad \quad ih429.cho@samsung.com}
\and
\IEEEauthorblockN{Cheul-Hee Hahm}
\IEEEauthorblockA{\textit{Samsung Electronics}\\
Suwon, Korea \\
chhahm@samsung.com}
}

\maketitle

\begin{abstract}
Video frame interpolation and prediction aim to synthesize frames in-between and subsequent to existing frames,
respectively. Despite being closely-related, these two tasks are traditionally studied with different model
architectures, or same architecture but individually trained weights. Furthermore, while arbitrary-time interpolation
has been extensively studied, the value of arbitrary-time prediction has been largely overlooked.  In this work, we
present \textit{uniVIP} - \textit{uni}fied arbitrary-time \textit{V}ideo \textit{I}nterpolation and \textit{P}rediction.
Technically, we firstly extend an interpolation-only network for arbitrary-time interpolation and prediction, with a
special input channel for task (interpolation or prediction) encoding.  Then, we show how to train a unified model on
common triplet frames. Our uniVIP provides competitive results for video interpolation, and outperforms existing
state-of-the-arts for video prediction. Codes will be available at: \url{https://github.com/srcn-ivl/uniVIP}
\end{abstract}


\section{Introduction}
\label{sec:intro}

Video frame interpolation is a long-standing vision task that aims to synthesize intermediate frames between consecutive
inputs. This technique can enhance video quality, and benefit downstream tasks like novel view
synthesis~\cite{flynn2016deepstereo} and video compression~\cite{lu2017novel}.  Video frame prediction aims to
synthesize frames subsequent to current ones. It can serve as a proxy task for representation
learning~\cite{oprea2020review}, and be applied in autonomous driving~\cite{castrejon2019improved} and human motion
prediction~\cite{martinez2017human}.

Both video interpolation~\cite{liu2017video,niklaus2020softmax,choi2020channel,park2021asymmetric,
huang2022rife,jia2022neighbor,jin2023unified} and
prediction~\cite{liu2017video,wu2022optimizing,jia2022neighbor,hu2023dynamic} have been widely studied in literature.
In particular, flow-based (\ie, warping-based) synthesis has developed into a common framework for both
tasks~\cite{liu2017video,huang2022rife,jia2022neighbor,jia2022neighbor,zhou2023exploring,hu2023dynamic}: optical flow is
firstly estimated to warp input frames, then warped frames are fused to synthesize the intermediate or next frame.
Besides similar technical framework, these two tasks also share similar key challenge, \ie, complex motion in in-the-wild
videos. While closely related, these two tasks are traditionally studied separately rather than unifiedly. Although a
few architectures can be applied for both tasks~\cite{liu2017video,jia2022neighbor,zhou2023exploring}, the model
weights have to be trained separately.

\begin{figure}[tb]
    \centering
    \includegraphics[width=1.0\columnwidth]{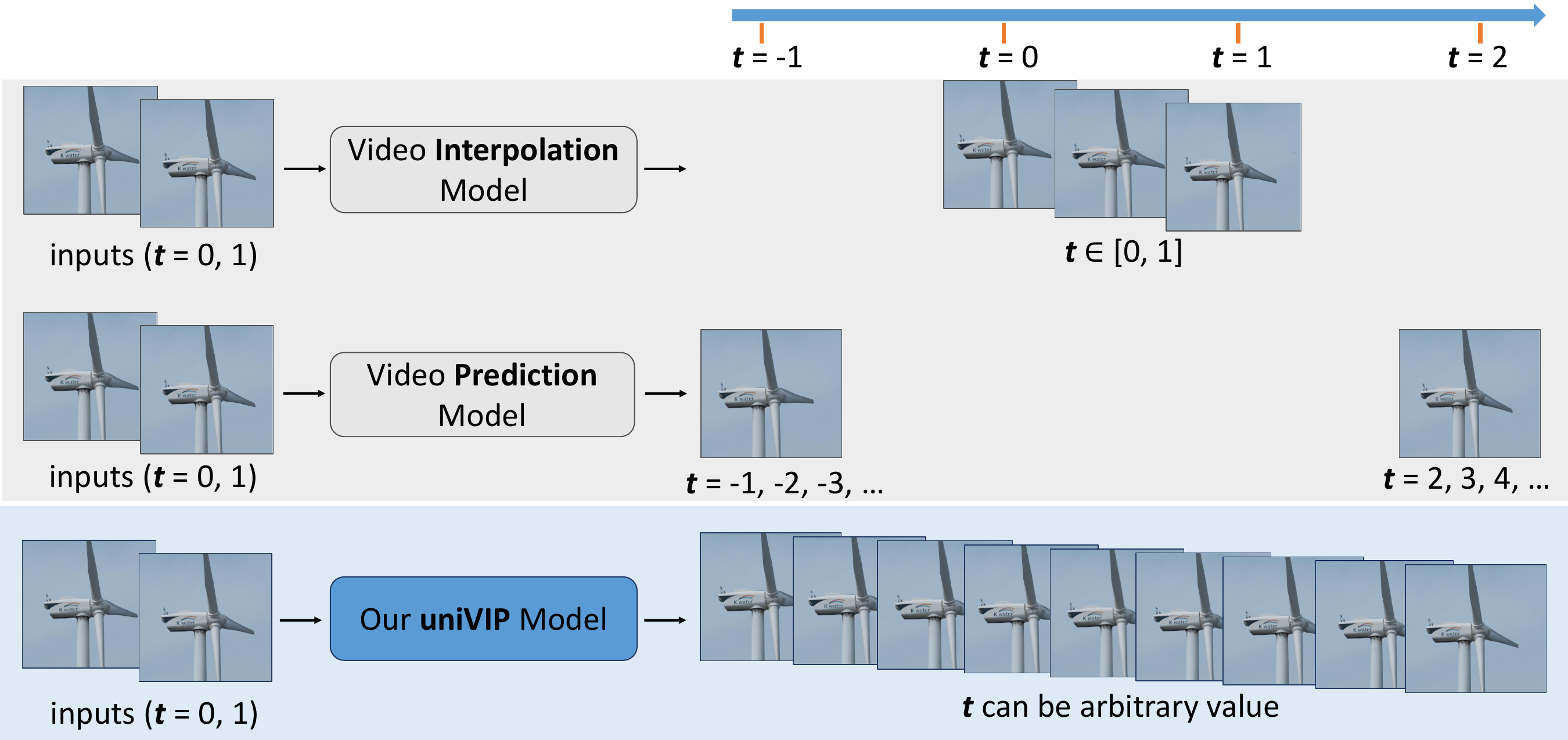}
    \caption{Conceptual comparisons between our uniVIP model and exiting video frame interpolation and prediction models. We
    pursue a unified model for both tasks, enabling frame synthesis at arbitrary-time (\eg, $t$ = 1.4 or -0.6).}
    \label{fig:intro}
\end{figure}

We ask: \textit{is it possible to train a \textbf{ single unified} model for \textbf{arbitrary-time} video interpolation
and prediction?} We pursue this goal due to two reasons. (\romannumeral1) A unified model can reduce deployment cost in
scenarios that involve both tasks, and may enjoy improved generalization via multi-task
learning~\cite{ghifary2015domain}.  (\romannumeral2) Existing video prediction methods cannot predict past and future
frames at arbitrary time, thus missing interesting applications like slow-motion of past or future frames.

In this work, we introduce uniVIP, unified arbitrary-time Video Interpolation and Prediction. As shown in
Fig.~\ref{fig:intro}, let $I_0$ and $I_1$ be input frames at time $t = 0$ and $t = 1$, our uniVIP can synthesize frames
at arbitrary time (\eg, $t$ = 1.4 or -0.6). By contrast, existing video interpolation methods can synthesize
arbitrary-time frames between input frames but cannot go
beyond~\cite{huang2022rife,park2021asymmetric,jin2023enhanced,jin2023unified}; existing video prediction methods can
only predict previous or future frames of integer indexes but cannot predict frames of floating indexes.

Given this, one might expect our uniVIP as a complicated and highly engineered method.  However, we show that uniVIP can
be easily derived, with simple adaptations to UPR-Net~\cite{jin2023unified}, an off-the-shelf forward-warping-based
interpolation model.  Specifically, we firstly extend UPR-Net for arbitrary-time video interpolation and prediction,
with a special channel for task (interpolation or prediction) encoding. Then, we show how to train an accurate unified
model for both tasks on common triplet frames~\cite{xue2019video}.

Our uniVIP, with a single model, can achieve high accuracy for both video interpolation and prediction. It provides
competitive results for video interpolation, and outperforms existing state-of-the-arts for video prediction.
Furthermore, to the best of our knowledge, our uniVIP is the first model that enables arbitrary-time frame prediction.

\section{Unified Arbitrary-time Video Frame Interpolation and Prediction}
\label{sec:approach}

\subsection{A Brief Recap of UPR-Net for Frame Interpolation}

Our uniVIP is built upon UPR-Net~\cite{jin2023unified}, a recent forward-warping-based video interpolation model.
UPR-Net iteratively refines both bi-directional flow and the intermediate frame within a recurrent pyramid framework. At
each pyramid level, UPR-Net follows the standard procedure of forward-warping-based interpolation: approximating optical
flow towards intermediate frame for forward warping; fusing warped input frames to synthesize intermediate frame.
UPR-Net is simple in design, excellent in performance, and enjoys the advantage of forward-warping for arbitrary-time
frame interpolation.  In this work, without causing ambiguity, we ignore the pyramid structure of UPR-Net to save space,
focusing on describing our adaptations to UPR-Net at each pyramid level.  We refer readers to~\cite{jin2023unified} for
the pyramid structure of UPR-Net.

\subsection{Designing uniVIP Model}

\subsubsection{Design Considerations and Overall Pipeline} 
\begin{figure}[tb]
\centering
\includegraphics[width=0.98\columnwidth]{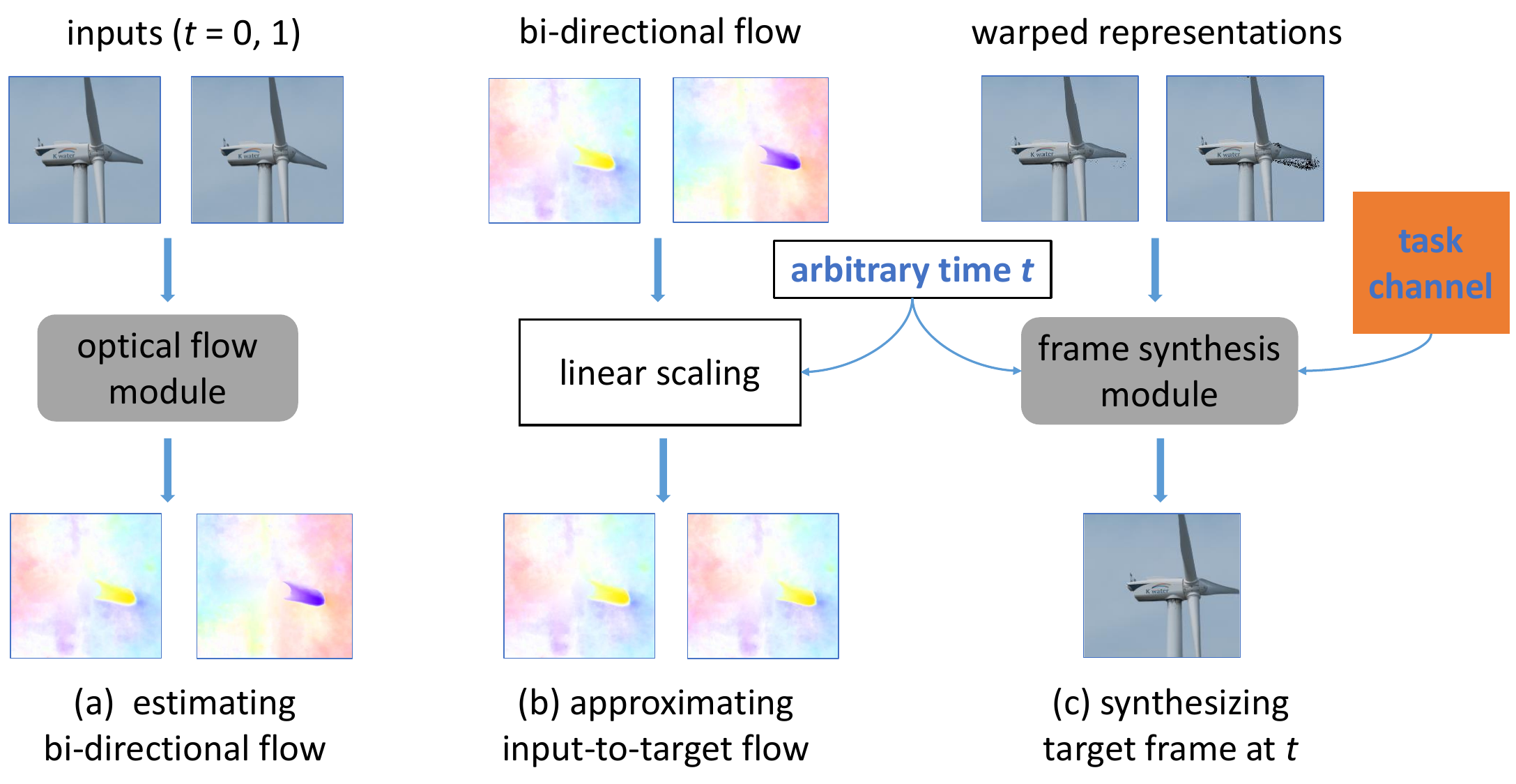}
\caption{Overview of uniVIP at each pyramid level (within a pyramid recurrent framework as in
UPR-Net~\cite{jin2023unified}). In our uniVIP, $t$ can be arbitrary value, with a special task channel for
differentiating interpolation and prediction tasks.}
\label{fig:architecture}
\end{figure}

Given $I_0$ and $I_1$, our goal is to synthesize $I_t$ at arbitrary time $t$: $t \in [0, 1]$ means interpolation; $t <
0$ or $t > 1$ means prediction.  For this goal, our adaptations to UPR-Net consist of three steps:
\begin{itemize}
    \item Estimating optical flow for forward warping, from input frames $I_0$ and $I_1$ to
        \textit{arbitrary-time} target frame $I_t$.
    \item Fusing forward-warped input frames to synthesize \textit{arbitrary-time} target frame $I_t$.
    \item Making synthesis module \textit{aware of task type}, since the warped frames for interpolation and
        prediction exhibit different patterns of artifact positions as elaborated later.
\end{itemize}
Fig.~\ref{fig:architecture} gives an overview of uniVIP. As uniVIP inherits the pyramid structure of UPR-Net, the
procedure in Fig.~\ref{fig:architecture} is repeated to refine the target frame from coarse to fine.

\subsubsection{Approximating Optical Flow for Arbitrary $t$}
With bi-directional flow $F_{0 \rightarrow 1}$ and $F_{1 \rightarrow 0}$ estimated by optical flow module, UPR-Net
linearly approximates $F_{0 \rightarrow t}$ and $F_{1 \rightarrow t}$ from $I_0$ and $I_1$ to \textit{intermediate}
$I_t$ (subjected to $t \in [0, 1]$):
{\small
\begin{equation}
    F_{0 \rightarrow t} = t \cdot F_{0 \rightarrow 1};  \\
    ~~~~~F_{1 \rightarrow t} = (1 - t) \cdot F_{1 \rightarrow 0}.
    \label{eq:linear}
\end{equation}
}%

We reveal that Eq.~\ref{eq:linear} also holds for \textit{both} $t < 0$ and $t > 1$.  For example, if $t < 0$, $F_{1
\rightarrow t} = (1 - t) \cdot F_{1 \rightarrow 0}$ holds, with $(1-t)$ increasing the magnitude of $F_{1 \rightarrow
0}$ but not changing the direction; $F_{0 \rightarrow t} = t \cdot F_{0 \rightarrow 1}$ also holds, with $t$ decreasing
the magnitude of $F_{0 \rightarrow 1}$, and reversing the flow direction. Despite being very simple, to the best of our
knowledge, Eq.~\ref{eq:linear} remaining valid for $t < 0$ and $t > 1$ has not been realized in previous video
interpolation and prediction works.

\subsubsection{Fusing Warped Frames to Synthesize $I_t$}
With $F_{0 \rightarrow t}$ and $F_{1 \rightarrow t}$, we can forward-warp input frames, and fuse warped frames to
synthesize $I_t$. We consider the target temporal location $t$ as a useful prior for temporal-dependent synthesis of
$I_t$, leading to following synthesis formulation:
{\small
\begin{align}\label{eq:synthesis}
    \hat{I_t} &= \frac{w_0 \cdot M_0 \odot I_{0 \rightarrow t} + w_1 \cdot M_1
        \odot I_{1 \rightarrow t}} {w_0 \cdot M_0 + w_1 \cdot M_1} + \Delta I_t  \nonumber\\
        w_0 & =
        \begin{cases}
        1 - t, ~~~~~~~~~~~~~~~~~~~~ if ~~~t \in [0, 1] \\  
        (1 - t) / (1 - 2 \cdot t), ~~~~ else
        \end{cases} \\
        w_1 & =
        \begin{cases}
        t, ~~~~~~~~~~~~~~~~~~~~~~~~~ if ~~~t \in [0, 1] \\  
        - t / (1 - 2 \cdot t), ~~~~~~~~~ else
        \end{cases} \nonumber
\end{align}
}%
where $M_0$ and $M_1$ are fusion maps for warped frames $I_{0 \rightarrow t}$ and $I_{1 \rightarrow t}$, $w_0$ and
$w_1$ are hard-coded temporal factors, and $\Delta I_t$ is the residual for further refinement. $M_0$ and
$M_1$ and $\Delta I_t$  are all estimated by the synthesis module. Note that Eq.~\ref{eq:synthesis} is inspired by
UPR-Net, but extends $t$ from $[0, 1]$ to arbitrary value in hard-coded temporal factors.

\subsubsection{Adding a Task Channel to Synthesis Module}
\begin{figure}[tb]
\centering
\includegraphics[width=0.95\columnwidth]{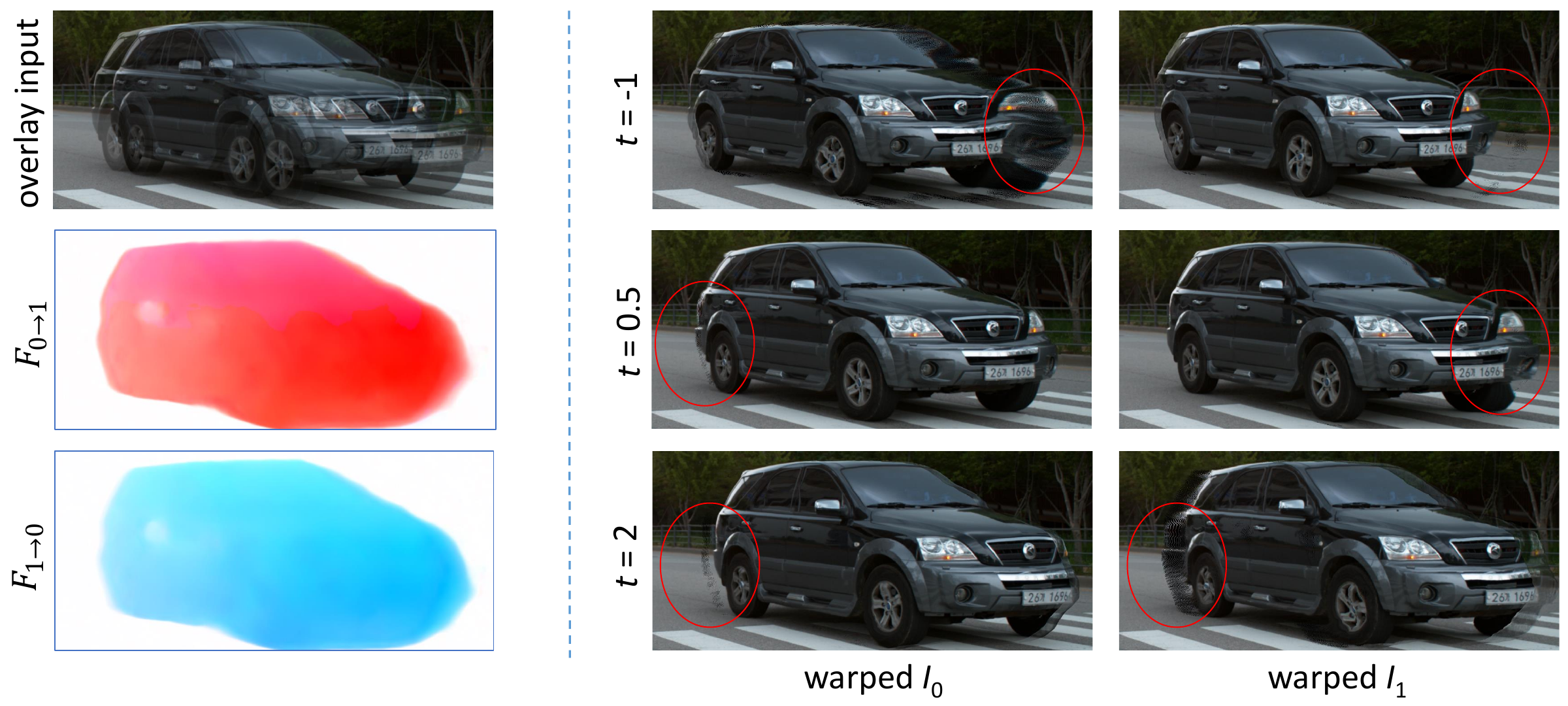}
\caption{For prediction (\eg, $t=2$), positions of artifacts in warped frames are similar, but for interpolation (\eg,
$t=0.5$), the positions are quite different.}
\label{fig:analysis}
\end{figure}

With above adaptations, it is already feasible for arbitrary-time interpolation and prediction.  However, for these
two tasks, we observe different patterns of artifact positions in warped frames\footnote{As explained
in~\cite{jin2023unified}, the artifacts themselves (\ie, holes by forward warping) in warped frames are inevitable in
case of large motion.}. As shown in Fig.~\ref{fig:analysis}, for video prediction, the artifacts appear in similar
positions, as optical flows from input frames to target frame are in the same direction.  While for video interpolation,
the artifacts typically appear in different positions because the flows are in opposite directions.

The different patterns of artifact positions raise difficulty and even ambiguity in training.  On one hand, the mapping
from warped frames to target frame should be task-dependent; on the other hand, the synthesis module, which should
remove the artifacts in warped frames, does not ``know'' current task type (interpolation or prediction).

To address this issue, we introduce a special task channel to ``notify'' the task type.  Specifically, this channel is
of the same size of warped images, and is filled with zeros for video interpolation, and ones for video prediction.  The
task channel allows the synthesis module to learn to handle task-specific artifact patterns in a data-driven manner.

\subsection{Training uniVIP Model}

\subsubsection{Training on Mixed Samples} 
We train uniVIP on mixed interpolation and prediction samples. Given three consecutive frames, when used for frame
interpolation, we denote them as $\{I_0, I_t, I_1\}$, and $t=0.5$ as it is the middle frame; when used for frame
prediction, we can construct two different samples: $\{I_{t}, I_0, I_1\}$ with $t = -1$ indicating previous
frame, and $\{I_0, I_1, I_t\}$ with $t = 2$ indicating next frame.
But, we only predict next frames during training, as explained below.

\subsubsection{Single-Directional Prediction} While uniVIP can predict both previous and future frames, 
single-directional prediction can make the training easier.  We predict future frames in training. While in testing, we
can convert previous frame prediction into future frame prediction.  Specifically, to predict $I_t$ with $t < 0$, we
reverse the order of input frames $I_0$ and $I_1$, switch their time step, and predict the frame at $(1 - t)$.

\subsection{Architecture Variants and Implementation Details}

\subsubsection{Architecture Variants}
We train two versions of uniVIP: (\romannumeral1) uniVIP-B, adapted from base version of
UPR-Net~\cite{jin2023unified}.  (\romannumeral2) uniVIP-L, adapted from LARGE version of
UPR-Net~\cite{jin2023unified}.

\subsubsection{Loss Function}
Our loss is the combined losses of video interpolation and prediction: $L = L_{interp} + L_{pred}$.
The reconstruction loss is the sum of Charbonnier loss~\cite{charbonnier1994two} and
census loss~\cite{meister2018unflow} between ground truth $I_t^{GT}$ and our
synthesized $I_{t}$: $\rho (I_t^{GT} - I_t) + L_{cen}(I_t^{GT}, I_t)$, where $\rho(x) = (x^2 +
\epsilon^2)^\alpha$ is the Charbonnier function, and $L_{cen}$ is the census loss.

\subsubsection{Training Dataset}
We use Vimeo90K dataset~\cite{xue2019video} for training, which contains 51,312 triplets with resolution of
$448\times256$ for training. We augment the training images by randomly cropping $256\times256$ patches.

\subsubsection{Optimization}
We use the AdamW~\cite{loshchilov2017decoupled} optimizer with weight decay $10^{-4}$ for 0.8 M iterations. The batch
size is set to  32. We gradually reduce the learning rate during training from $2\times10^{-4}$ to $2\times10^{-5}$
using cosine annealing.

\section{Experiments}
\label{sec:exp}

\subsection{Evaluation Settings}
\label{exp:setting}

\subsubsection{Evaluation Datasets}
We evaluate our models on below benchmarks for \textit{both} interpolation and prediction.
\begin{itemize}
    \item Vimeo90K~\cite{xue2019video}. The test set of Vimeo90K contains 3,782 triplets, with a resolution of
        $448\times256$.
    \item SNU-FILM~\cite{choi2020channel}: This dataset contains 1,240 triplets, with resolution around
        1280$\times$720. It contains four subsets -- easy, medium, hard, and extreme.
    \item X-TEST~\cite{sim2021xvfi}: It is a high-quality benchmark of 4K resolution, originally proposed for $\times$8
        frame interpolation.
\end{itemize}

\subsubsection{Dataset Configurations}
On Vimeo90K and SNU-FILM, we evaluate middle-frame interpolation and next-frame prediction.  While for X-TEST, we use
it for three purposes: $\times$8 frame interpolation, next-frame prediction, and arbitrary-time frame prediction.
Specifically, we simulate next-frame prediction ($t=2.0$) by predicting the 32-nd frame based on the 0-th and the 16-th
frames. For arbitrary-time prediction, the configuration will be detailed later in context.

\subsubsection{Evaluation Metrics}
We measure the peak signal-to-noise ratio (PSNR) and structure similarity (SSIM)~\cite{wang2004image} for quantitative
evaluation. For running time, we test our models with a RTX 2080 Ti GPU for interpolating $640\times480$ inputs.

\subsection{Video Frame Interpolation and Next-Frame Prediction}

\begin{table}[tb]
\scriptsize
\centering
\setlength{\tabcolsep}{0pt}
\begin{tabular*}{1.0\columnwidth}{@{\extracolsep{\fill}}*{8}{lcccccc}}
\hline
\multirow{2}{*}{Methods} & \multirow{2}{*}{Vimeo90K} & \multicolumn{4}{c}{SNU-FILM} & \multirow{2}{*}{X-TEST}\\
\cline{3-6}
                        &  &  Easy & Medium & Hard & Extreme & \\
\Xhline{2\arrayrulewidth}
\multicolumn{7}{l}{\textit{$\times$2 (Vimeo90K, SNU-FILM) and $\times$8 (X-TEST) Frame Interpolation}}   \\
RIFE & 35.61/.9779 & 40.06/.9905 & 35.73/.9787 
                          & 30.08/.9328 & 24.82/.8530 & 29.14/.8765\\
RIFE-L & 36.13/.9800 & 40.23/.9907 & 35.86/.9792 & 30.19/.9332 & 24.81/.8540 & 28.94/.8721\\
VFIformer &\textcolor{red}{36.50}/\textcolor{red}{.9816} 
          & 40.13/.9907 & 36.09/\textcolor{blue}{\underline{.9799}} & 30.67/\textcolor{red}{.9378}
          & 25.43/.8643 & - \\
NCM-B     & 35.88/.9795 & 39.98/.9903 & 35.94/.9788 & 30.72/.9359 & 25.55/.8624 
                                 & \textcolor{blue}{\underline{31.63}}/\textcolor{blue}{\underline{.9185}}\\
NCM-L     & 36.22/.9807 & 40.14/.9905 & 36.12/.9793
                                 & \textcolor{red}{30.88}/.9370
                                 & \textcolor{red}{25.70}/\textcolor{blue}{\underline{.8647}} 
                                 & \textcolor{red}{31.86}/\textcolor{red}{.9225}\\
UPR-Net-L  &\textcolor{blue}{\underline{36.42}}/\textcolor{blue}{\underline{.9815}}
                  &\textcolor{blue}{\underline{40.44}}/\textcolor{red}{.9911}
                  &\textcolor{red}{36.29}/\textcolor{red}{.9801}
                  &\textcolor{blue}{\underline{30.86}}/\textcolor{red}{.9377}
                  &\textcolor{blue}{\underline{25.63}}/.8641 & 30.50/.9048\\
\textit{uniVIP-B}  & 35.87/.9796 & 40.33/\textcolor{blue}{\underline{.9910}} & 36.07/.9795 & 30.62/.9346 & 25.50/.8635
                  & 31.05/.9130\\
\textit{uniVIP-L} & 36.26/.9810
                  & \textcolor{red}{40.49}/\textcolor{red}{.9911} 
                  & \textcolor{blue}{\underline{36.26}}/\textcolor{red}{.9801} 
                  & 30.78/.9373 & 25.61/\textcolor{red}{.8648} 
                  & 31.03/.9163\\
\Xhline{2\arrayrulewidth}
\multicolumn{7}{l}{\textit{Next Frame Prediction (input $I_0$ and $I_1$, predict $I_2$)}}    \\
RIFE         & 31.70/.9583 & 36.72/.9822 & 32.35/.9568
                                  & 27.17/.8883 & 22.63/.8061 & 23.36/.7845 \\
RIFE-L      & 31.99/.9609 & 36.59/.9820 
                                 & 32.27/.9567 & 27.14/.8890 & 22.60/.8075  & 23.10/.7843\\
NCM-B     & 31.84/.9595 & 36.46/.9814 & 32.24/.9557 & 27.28/.8881 
                                 & 22.84/.8090  & 27.69/.8694\\
NCM-L     & \textcolor{blue}{\underline{32.10}}/\textcolor{blue}{\underline{.9614}}
                                   & 36.40/.9814 & 32.24/.9561
                                   & \textcolor{blue}{\underline{27.41}}/.8895
                                   & \textcolor{red}{23.00}/.8120 
                                   & \textcolor{blue}{\underline{28.05}}/\textcolor{blue}{\underline{.8786}}\\
DMVFN        & 30.31/.9467 & 36.10/.9802 & 31.57/.9526 & 26.39/.8804 
                                  & 21.93/.7925 & 22.50/.7668 \\
\textit{uniVIP-B}  & 32.01/.9611
                   & \textcolor{blue}{\underline{36.87}}/\textcolor{red}{.9823}
                   & \textcolor{blue}{\underline{32.47}}/\textcolor{blue}{\underline{.9571}}
                   & 27.40/\textcolor{blue}{\underline{.8902}} 
                   & \textcolor{blue}{\underline{22.95}}/\textcolor{blue}{\underline{.8124}} 
                   & 28.00/.8776\\
\textit{uniVIP-L}  & \textcolor{red}{32.20}/\textcolor{red}{.9625}
                  & \textcolor{red}{36.89}/\textcolor{red}{.9824}
                  & \textcolor{red}{32.55}/\textcolor{red}{.9576}
                  & \textcolor{red}{27.45}/\textcolor{red}{.8909}
                  & \textcolor{red}{23.00}/\textcolor{red}{.8132} 
                  & \textcolor{red}{28.35}/\textcolor{red}{.8810}\\
\hline
\end{tabular*}
\caption{Qualitative (PSNR/SSIM) comparisons of video interpolation and prediction. \textcolor{red}{RED}: best
performance, \textcolor{blue}{\underline{BLUE}}: second best.}
\label{tab:three-benmarks}
\end{table}

\subsubsection{Video Frame Interpolation}
As shown in Tab.~\ref{tab:three-benmarks}, our uniVIP-L provides competitive results with state-of-the-art
interpolation methods including RIFE~\cite{huang2022rife}, VFIformer~\cite{lu2022video}, UPR-Net~\cite{jin2023unified},
and NCM~\cite{jia2022neighbor}.
When trained for both interpolation and prediction, it achieves similar interpolation accuracy with UPR-Net-L which is
trained only for video interpolation.  For $\times$8 interpolation on X-TEST, uniVIP significantly outperforms RIFE and
UPR-Net-L, but is inferior to NCM which is specially designed for high-resolution synthesis.

\subsubsection{Next-Frame Prediction}

Tab.~\ref{tab:three-benmarks} also shows the comparison results for next frame prediction.
For fair comparison, we re-train RIFE~\cite{huang2022rife} and DMVFN~\cite{hu2023dynamic} on Vimeo90K for prediction.
Our uniVIP-L sets new state-of-the-art on most of these benchmarks. Our uniVIP-B model, with much less
parameters than NCM (1.7M VS. 12.1M), also outperforms NCM-L which is designed for high-resolution synthesis.

\begin{figure*}[h]
  \centering
  \begin{minipage}{0.45\columnwidth}
    \centering
    \includegraphics[width=\columnwidth]{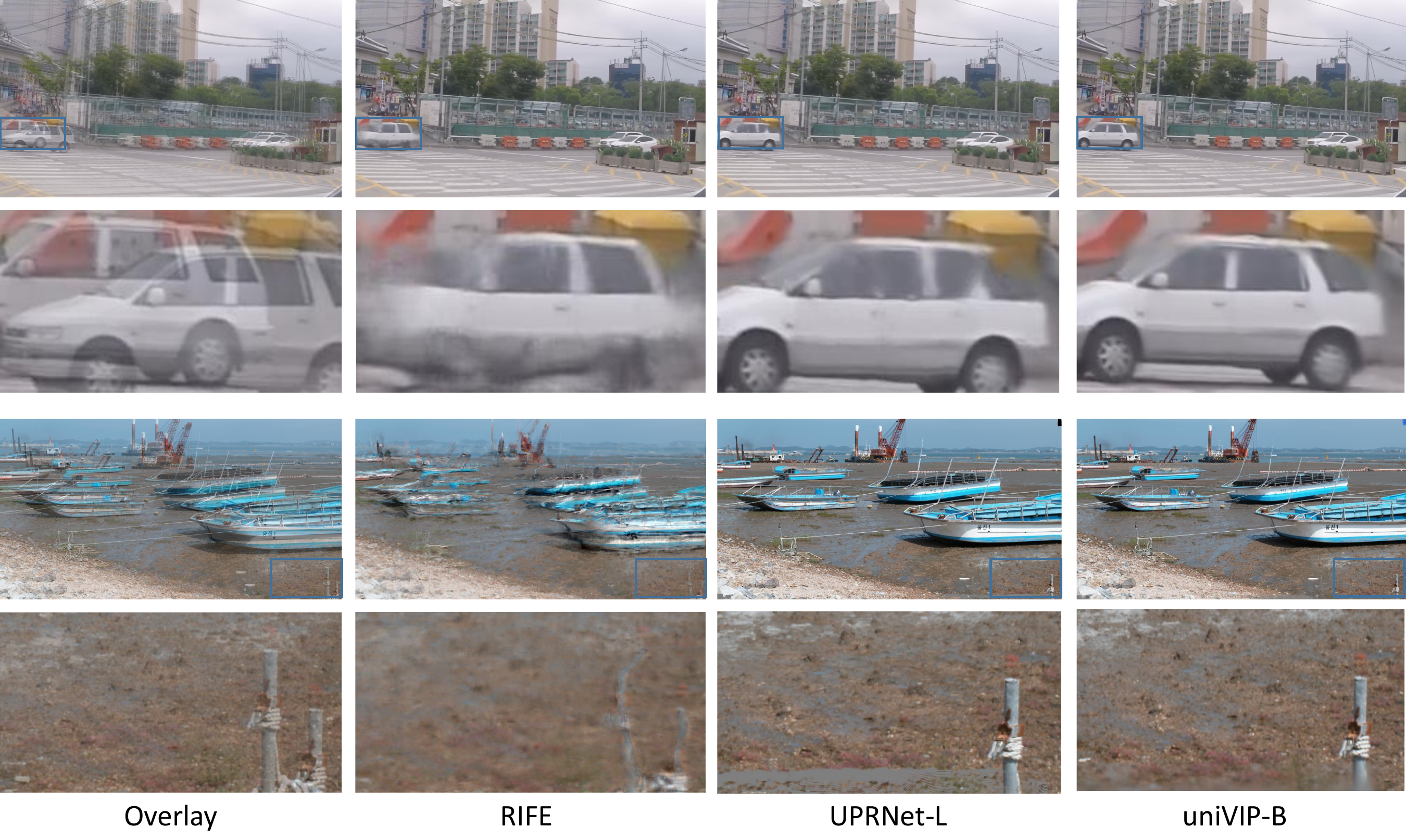}
  \end{minipage}
  \hfill
  \begin{minipage}{0.45\columnwidth}
    \centering
    \includegraphics[width=\columnwidth]{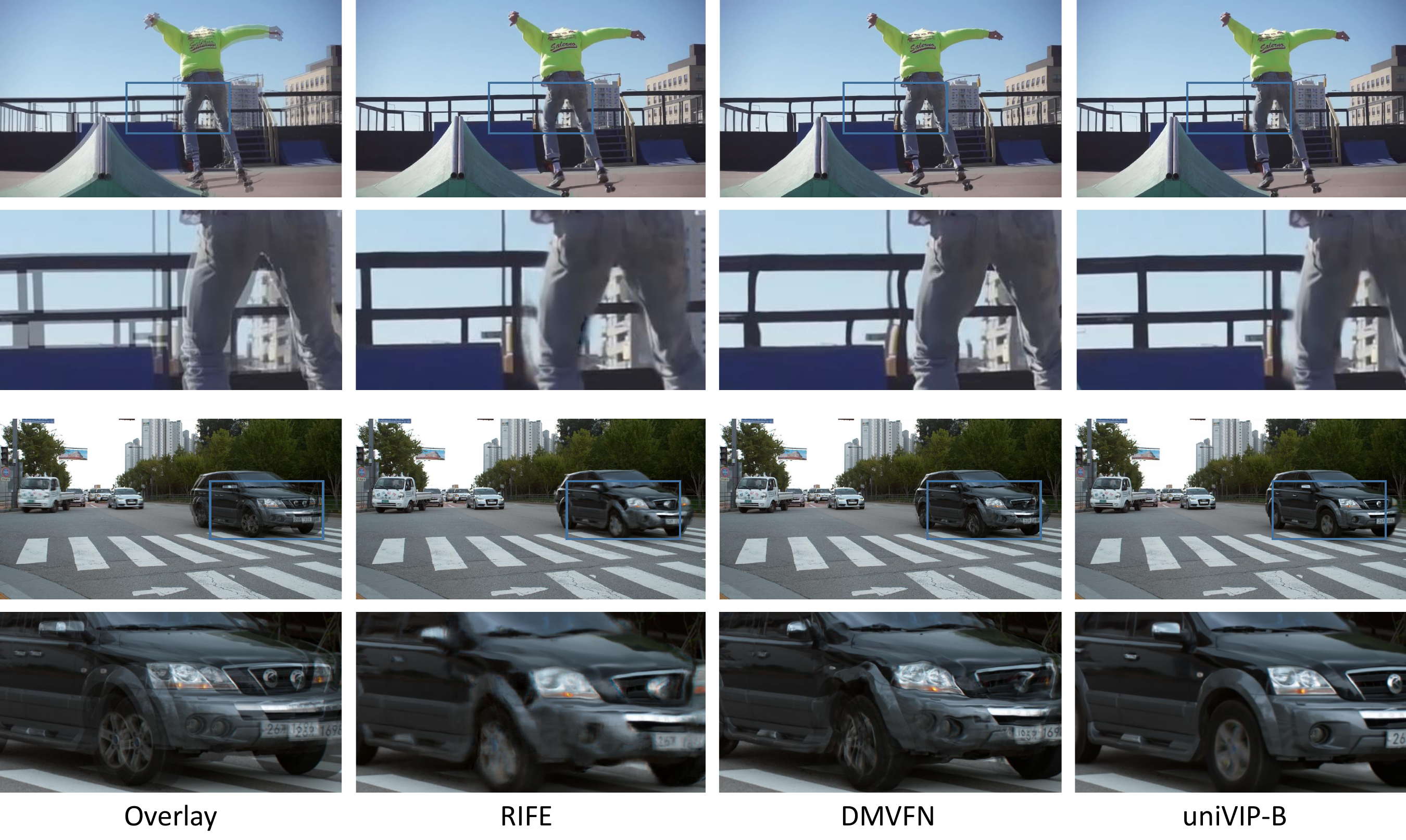}
  \end{minipage}
  \caption{\textbf{Left:} examples from SNU-FILM and X-TEST for interpolation. \textbf{Right:} examples from SNU-FILM
  and X-TEST for prediction.}
  \label{fig:visualization}
\end{figure*}

\subsubsection{Visual Comparisons}
Fig.~\ref{fig:visualization} shows typical examples from SNU-FILM~\cite{choi2020channel} and X-TEST~\cite{sim2021xvfi}
for video interpolation and prediction.  Our uniVIP is robust to extreme large motion, with plausible synthesis of the
regions close to image border which may suffer from serious artifacts in warped frames.

\subsection{Arbitrary-Time Video Frame Prediction}

\begin{table}[tb]
\centering
\scriptsize
\setlength{\tabcolsep}{0pt}
\begin{tabular*}{1.0\columnwidth}{@{\extracolsep{\fill}}*{15}{lccccccccccccc}}
\hline
Methods & \multicolumn{13}{c}{Arbitrary-time Frame Prediction on X-TEST} \\
\Xhline{2\arrayrulewidth}
        & \multicolumn{13}{c}{\textit{Time Step $t$ in Predicting \textbf{Previous} Frames ($t < 0$)}}   \\
& -0.25 & -0.50 & -0.75
             & -1.00 & -1.25 & -1.50 & -1.75 & -2.00 
             & -2.25 & -2.50 & -2.75 & -3.00  & avg.\\
\cline{2-14}
RIFE  & - & - & - & 28.81 & - & - & - & 25.95 & - & - & -  & 24.01 & 26.26 \\
RIFE-L & - & - & - & 28.40 & - & - & -  & 25.79 & - & - & - & 23.92  & 26.04 \\
DMVFN   & - & - &  - & 27.01 & - & - & - & 24.47 & - & - & - & 22.71 & 24.73 \\
\textit{uniVIP-B} & \textcolor{red}{35.74} & \textcolor{blue}{\underline{34.35}}         
         & \textcolor{blue}{\underline{33.50}} & \textcolor{blue}{\underline{31.96}}
         & \textcolor{blue}{\underline{30.80}}
         & \textcolor{blue}{\underline{29.46}}
         & \textcolor{blue}{\underline{28.58}}
         & \textcolor{blue}{\underline{27.59}}
         & \textcolor{blue}{\underline{26.95}}
         & \textcolor{blue}{\underline{26.20}}
         & \textcolor{blue}{\underline{25.69}}
         & \textcolor{blue}{\underline{25.09}}
         & \textcolor{blue}{\underline{29.66}} \\
\textit{uniVIP-L} & \textcolor{blue}{\underline{35.22}} & \textcolor{red}{34.38}
         & \textcolor{red}{33.79} & \textcolor{red}{32.37}
         & \textcolor{red}{31.19} & \textcolor{red}{29.74}
         & \textcolor{red}{28.82}
         & \textcolor{red}{27.79}
         & \textcolor{red}{27.12}
         & \textcolor{red}{26.33}
         & \textcolor{red}{25.80}
         & \textcolor{red}{25.16}
         & \textcolor{red}{29.81}  \\
\hline
        & \multicolumn{13}{c}{\textit{Time Step $t$ in Predicting \textbf{Future} Frames ($t > 1$)}}   \\
& 1.25 & 1.50 & 1.75
             & 2.00 & 2.25 & 2.50 & 2.75 & 3.00 
             & 3.25 & 3.50 & 3.75 & 4.00 & avg.\\
\cline{2-14}
RIFE  & - & - & - & 28.81 & - & - & - & 26.12 & - & - & -  & 24.37 & 26.43\\
RIFE-L & - & - & - & 28.27 & - & - & -  & 25.73 & - & - & - & 24.05 & 26.02\\
DMVFN   & - & - &  - & 27.48 & - & - & - & 24.72 & - & - & - & 22.90 & 25.03\\
\textit{uniVIP-B} & \color{red}{35.64} & \textcolor{blue}{\underline{34.44}}
         & \textcolor{blue}{\underline{33.62}} & \textcolor{blue}{\underline{32.06}}
         & \textcolor{blue}{\underline{30.92}}
         & \textcolor{blue}{\underline{29.65}}
         & \textcolor{red}{28.77}
         & \textcolor{red}{27.84}
         & \textcolor{red}{27.18}
         & \textcolor{red}{26.47}
         & \textcolor{red}{25.94}
         & \textcolor{red}{25.39} 
         & \textcolor{blue}{\underline{29.83}} \\
\textit{uniVIP-L} & \color{blue}{\underline{35.29}} & \textcolor{red}{34.61}
         & \textcolor{red}{34.03} & \textcolor{red}{32.39}
         & \textcolor{red}{31.08}
         & \textcolor{red}{29.70}
         & \textcolor{red}{28.77}
         & \textcolor{blue}{\underline{27.79}}
         & \textcolor{blue}{\underline{27.09}} 
         & \textcolor{blue}{\underline{26.33}}
         & \textcolor{blue}{\underline{25.78}} 
         & \textcolor{blue}{\underline{25.22}}  
         & \textcolor{red}{29.84} \\
\hline
\end{tabular*}
\caption{Arbitrary-time frame prediction on X-TEST~\cite{xue2019video}. \textcolor{red}{RED}: best performance,
\textcolor{blue}{\underline{BLUE}}: second best performance.}
\label{tab:x-test}
\end{table}

We evaluate uniVIP for arbitrary-time prediction on X-TEST.  In particular, when predicting previous frames ($t<0$), we
choose 24-th and 32-th frames as inputs; as for future frames, we choose 0-th and 8-the frames as inputs.

As in Tab.~\ref{tab:x-test}, our uniVIP significantly outperforms RIFE, RIFE-L and DMVFN which can only predict frames
of integer indexes. Furthermore, our performance of predicting frames of floating indexes is similar to integer-indexed
prediction.

\subsection{Comparisons on Flexibility, Parameters, and Runtime}

\begin{table}[tb]
\centering
\scriptsize
\setlength{\tabcolsep}{0pt}
\begin{tabular*}{0.8\columnwidth}{@{\extracolsep{\fill}}*{7}{lccccc}}
\hline

\multirow{2}{*}{Methods} & \multirow{2}{*}{unified model}
                         & \multicolumn{2}{c}{arbitrary-time} 
                         &  parameters & runtime \\
\cline{3-4}
                         & & interpolation & prediction  & (millions) & (seconds) \\

\Xhline{2\arrayrulewidth}
RIFE$_m$ &  $\times$ & \checkmark & unclear  & 10.1 & 0.012   \\
RIFE &  $\times$  & $\times$ & $\times$  & 10.1 & 0.012   \\
RIFE-L   &  $\times$  & $\times$ & $\times$  & 10.1 & 0.068  \\
NCM-B    & $\times$ & $\times$ & $\times$ & 12.1 & 0.038  \\
NCM-L    & $\times$ & $\times$ & $\times$ & 12.1 & 0.122  \\
\textit{uniVIP-B} &\checkmark & \checkmark & \checkmark
                  & 1.7 & 0.042  \\
\textit{uniVIP-L} &\checkmark & \checkmark & \checkmark
                  & 6.6 & 0.081  \\
\hline
\end{tabular*}
\caption{Comparisons in architecture, parameters, and runtime.}
\label{tab:param-runtime}
\end{table}

In Tab.~\ref{tab:param-runtime}, we compare uniVIP with RIFE and NCM on architecture, parameters and runtime, where
uniVIP shows advantages in unified modeling and arbitrary-time prediction.

\subsection{Ablation Experiments}
\label{exp:ablation}

\begin{table}[tb]
\scriptsize
\centering
\setlength{\tabcolsep}{0pt}
\begin{tabular*}{1.0\columnwidth}{@{\extracolsep{\fill}}*{11}{cccccccccc}}
\hline
\multirow{3}{*}{Experiments} & \multirow{3}{*}{Methods} 
& \multicolumn{3}{c}{Extreme} & \multicolumn{5}{c}{X-TEST} \\
\cline{3-5}
\cline{6-10}
& & $\times$2 &  $\times$1 & $\times$1 
& $\times$8 &  $\times$8 &  $\times$8 &  $\times$24 &  $\times$24 \\
& & interp. &  prev. & future & interp. &  prev. &  future &  prev. &  future\\
\Xhline{2\arrayrulewidth}
\multirow{2}{*}{temporal factor} & without & 25.51 & 23.05 & 22.96 & 30.41 
                                 & 29.60 & 29.80 & 30.90 & 30.95 \\ 
                                   & with & 25.50 & 23.05
                                   & 22.95 & \textbf{31.05} & \textbf{29.66} & 29.83 & \textbf{31.00} & \textbf{31.09} \\
\hline
\multirow{2}{*}{convt. prev-pred.} & direct & 25.47 & 23.02 & 22.93 & 30.95 
                                  & 29.55 & 29.87 & 31.00 & 30.98 \\
                               & convt. & 25.50 & 23.05 
                                   & 22.95 & \textbf{31.05} & \textbf{29.66} & 29.83 & 31.00 & \textbf{31.09} \\
\hline
\multirow{2}{*}{task channel} & without & 25.49 & 22.89 & 22.82 & 30.40 & 29.30 & 29.35 & 31.02 & 31.01 \\
                               & with & 25.50 & \textbf{23.05}
                               & \textbf{22.95} & \textbf{31.05} & \textbf{29.66} & \textbf{29.83} & 31.00 &
                               \textbf{31.09}\\
\hline
\multirow{2}{*}{unified model} & individual & 25.51 & 23.03 & 22.94 & 31.08 
                               & 29.45 & 29.54 & 28.45 & 28.68 \\
                               & unified & 25.50 & 23.05 
                               & 22.95 & 31.05 & \textbf{29.66} & \textbf{29.83} & \textbf{31.00} & \textbf{31.09}\\
\hline
\end{tabular*}
\caption{Ablation experiments measured by PSNR}
\label{tab:ablation}
\end{table}

We conduct ablation experiments on SNU-FILM extreme subset and X-TEST.  On X-TEST, we choose 0-th and 8-th frames as
input for $\times$8 future prediction; and 0-th and 24-th for $\times$24 future prediction.  We use uniVIP-B for all
experiments, and summarize all results in Tab.~\ref{tab:ablation}.

\paragraph{Temporal Factor for Fusion}
Temporal factor in Eq.\ref{eq:synthesis} has little impact on moderate-resolution extreme subset, but significantly
increases the accuracy of $\times$8 interpolation, and slightly improves multiple frame prediction on X-TEST.

\paragraph{Converted Previous Frame Prediction}
We verify the effectiveness of single-direction prediction in training, with comparison to both previous and next
prediction in training.

\paragraph{Task Channel for Unified Frame Synthesis}
Our special task channel to the synthesis module brings obvious improvements on challenging X-TEST dataset. 

\paragraph{Individual Models \textit{VS}. uniVIP}
Comapred to individually trained interpolation and prediction models, our uniVIP produces competitive results on extreme
subset, and much better accuracy on X-TEST for prediction.

\section{Conclusion}
\label{sec:conclusion}
We presented uniVIP for arbitrary-time video interpolation and prediction. Despite its simple design, uniVIP achieved
excellent accuracy for both tasks, and enabled arbitrary-time video frame prediction.
In the future, we will seek to apply uniVIP on various real-life applications.

\bibliographystyle{IEEEtran}
\bibliography{univip}

\end{document}